\documentclass[letterpaper, 10 pt, journal, twoside]{IEEEtran}
\usepackage{graphicx} %
\graphicspath{ {03_figures/} }
\usepackage{hyperref}

\usepackage[nolist,nohyperlinks]{acronym}

\usepackage{balance}
\usepackage[backend=bibtex, style=ieee, isbn=false, doi=false, url=false]{biblatex}
\AtBeginBibliography{\small}

\usepackage{mathtools}%

\makeatletter
\let\MYcaption\@makecaption
\makeatother

\usepackage[font=footnotesize]{subcaption}

\usepackage{microtype}

\makeatletter
\let\@makecaption\MYcaption
\makeatother

\usepackage{enumerate}%

\usepackage{arydshln}       %

\usepackage{tikz}

\usepackage[commandnameprefix=always]{changes}

\usepackage{setspace}       %

\bibliography{02_resources/library_nicolai, 02_resources/library}

\usepackage{eso-pic}
\usepackage{url}
\AddToShipoutPicture*{%
     \AtTextUpperLeft{%
         \put(-3.5,10){
           \begin{minipage}{\textwidth}
              \scriptsize
              \MakeUppercase{Preprint version of a 2025 Neuro Inspired Computational Elements Conference (NICE) proceedings paper\\Final version available at} \url{https://ieeexplore.ieee.org/}
           \end{minipage}}%
     }%
}

\begin{document}

\title{Threshold Adaptation in Spiking Networks Enables Shortest Path Finding and Place Disambiguation}

\author{
    Robin Dietrich$^{1,2*}$\hskip3em Tobias Fischer$^2$\hskip3em Nicolai Waniek$^3$\hskip3em Nico Reeb$^{1,2}$\hskip3em Michael Milford$^2$\hskip3em Alois Knoll$^{1}$\hskip3em Adam D. Hines$^2$ \\
    \thanks{
        This research was partially supported by funding from ARC DECRA Fellowship DE240100149 to TF and an ARC Laureate Fellowship FL210100156 to MM. The authors acknowledge continued support from the Queensland University of Technology (QUT) through the Centre for Robotics. This work has been supported by a fellowship of the German Academic Exchange Service (DAAD) and by the FFG, Contract No. 881844: ``Pro$^2$Future''. 
        \vspace{5pt}
        \\
        $^1$School of Computation, Information and Technology, Technical University of Munich, Munich, Germany. \\
        $^2$QUT Centre for Robotics, School of Electrical Engineering and Robotics, Queensland University of Technology, Brisbane, Australia. \\
        $^3$Department of Mathematical Sciences, Norwegian University of Science and Technology, Trondheim, Norway. \\
        $^*$robin.dietrich@tum.de
    }
}

\maketitle

\thispagestyle{empty}

\newcommand{\subpop}[1]{\mathcal{M}_{#1}}
\newcommand{\subpopm}[1]{$\mathcal{M}_{#1}$}

\newcommand{\loc}[1]{\sym{#1}}
\newcommand{\sym}[1]{S_{#1}}

\newcommand{\tmaxb}[0]{\Delta t_{\text{max}}^{b}}
\begin{abstract}

Efficient spatial navigation is a hallmark of the mammalian brain, inspiring the development of neuromorphic systems that mimic biological principles. Despite progress, implementing key operations like back-tracing and handling ambiguity in bio-inspired spiking neural networks remains an open challenge. 
This work proposes a mechanism for activity back-tracing in arbitrary, uni-directional spiking neuron graphs. We extend the existing replay mechanism of the \acf{shtm} by our \acf{stdta}, which enables us to perform path planning in networks of spiking neurons. We further present an \acf{adta} for identifying places in an environment with less ambiguity, enhancing the localization estimate of an agent. Combined, these methods enable efficient identification of the shortest path to an unambiguous target. 
Our experiments show that a network trained on sequences reliably computes shortest paths with fewer replays than the steps required to reach the target. We further show that we can identify places with reduced ambiguity in multiple, similar environments. These contributions advance the practical application of biologically inspired sequential learning algorithms like the \ac{shtm} towards neuromorphic localization and navigation.

\vspace{-5pt}
\end{abstract}

\begin{acronym}

\acro{mc}[MC]{multi-compartment}
\acro{snn}[SNN]{spiking neural network}
\acro{ann}[ANN]{artificial neural network}
\acro{htm}[HTM]{hierarchical temporal memory}
\acro{shtm}[S-HTM]{spiking hierarchical temporal memory}
\acro{sdr}[SDR]{sparse distributed representation}
\acro{diw}[DIW]{dynamic excitatory-inhibitory weights}

\acro{stdp}[STDP]{spike timing-dependent plasticity}
\acro{sstdp}[sSTDP]{structural spike timing-dependent plasticity}
\acro{stdta}[STDTA]{spike timing-dependent threshold adaptation}
\acro{adta}[ADTA]{ambiguity dependent threshold adaptation}
\acro{nmda}[NMDA]{N-Methyl-D-Aspartat}
\acro{dap}[dAP]{dendritic action potential}
\acro{bt}[BT]{back-tracing}

\acro{bss}[BSS-2]{\mbox{BrainScaleS-2}}
\acro{fpga}[FPGA]{Field Programmable Gate Array}
\acro{isa}[ISA]{instruction set architecture}
\acro{simd}[SIMD]{single instruction, multiple data}
\acro{adc}[ADC]{analog-to-digital converter}

\end{acronym}

\section{Introduction}

Localization is an important process of many navigation systems, providing the ability to know where in the world you are based on a variety of sensory modalities~\cite{borenstein1996navigatingrobots,burgard2011MarkovLocalizationMobile}.
Modern techniques that perform localization often rely on supervised deep-learning architectures, which require manually labeled datasets that can require high energy consumption, with long training and deployment times~\cite{mokssit2023DeepLearningTechniques}. Therefore, there is significant interest in exploring alternative computing architectures that can be used for real-world scenarios in robotic deployment on size, weight and power (SWaP) constrained platforms.

A promising direction for such navigation systems is neuromorphic computing, which is inspired by the dynamics of biological neurons. Neuromorphic systems have been shown to perform low-power, low-latency operations in many applications~\cite{paredesValls2024fullyneuromorphic,Stroobants2023neuromorphiccontrol,tang2019spiking,harbour2024martianflight,oess2020NearOptimalBayesianIntegration,perez-nieves2022SparseSpikingGradient}. Neuromorphic systems are highly specialized accelerators for spiking neural networks (SNNs), which operate on the principle of massively parallel, asynchronous, sparse data processing and representations to achieve energy and computational efficiency~\cite{goltz2021FastEnergyefficientNeuromorphic,barnell2023ultralowpower,kadway2023LowPowerLow}. Nonetheless, the development of neuromorphic computing faces important open questions, particularly in the selection of neuron models and learning algorithms that optimally utilize the diverse set of existing neuromorphic hardware~\cite{furber2014SpiNNakerProjecta,pei2019ArtificialGeneralIntelligence,davies2018LoihiNeuromorphicManycore,richter2024DYNAPSE2ScalableMulticore,kadway2023LowPowerLow,akopyan2015TrueNorthDesignTool}.

\begin{figure}[!t]
  \centering
  \includegraphics[width=0.9\columnwidth]{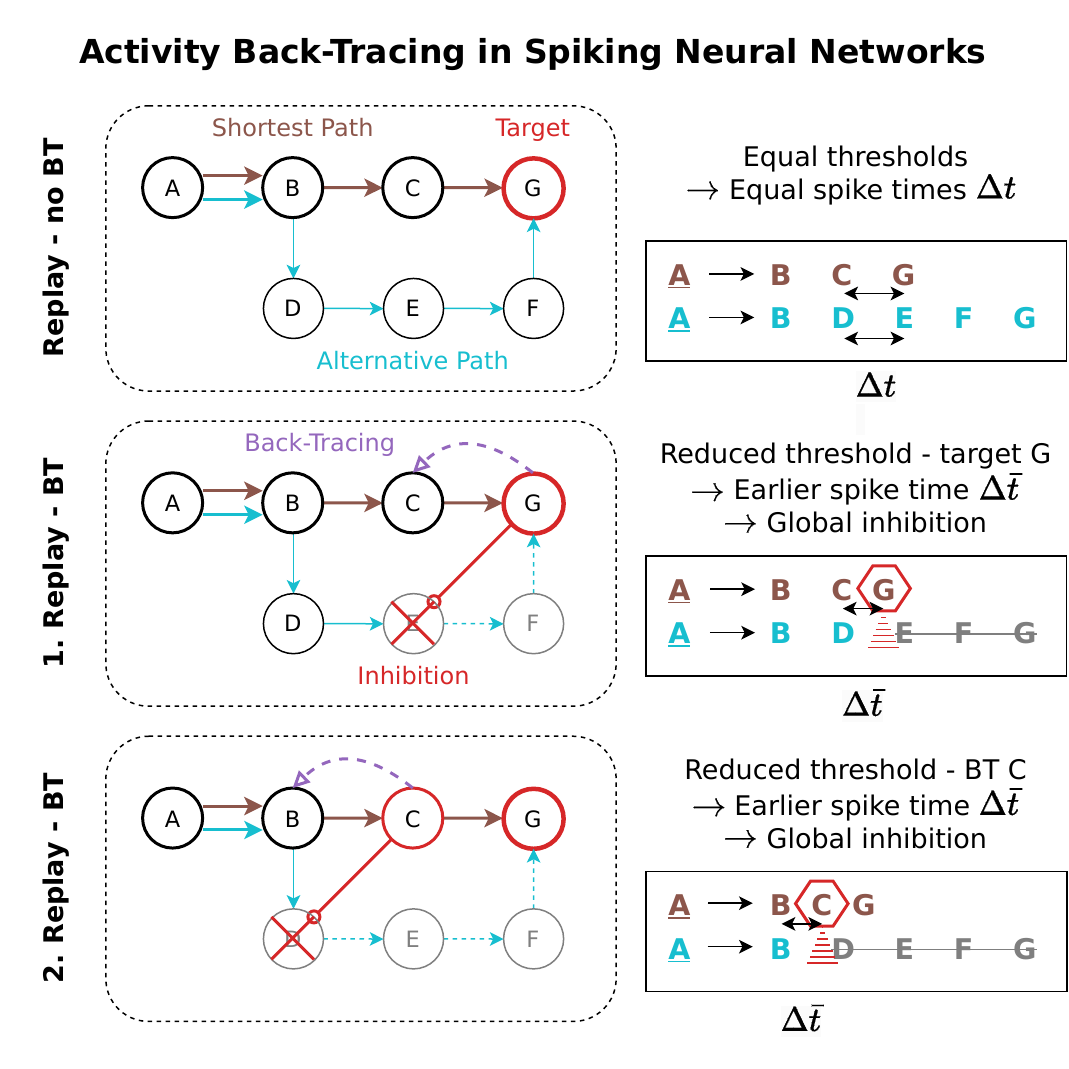}
  \caption{(Left) The uni-directionally connected network graph during replay, initiated at \textit{A} and with two paths to the target \textit{G}. Shown are a replay without back-tracing and two replays with back-tracing, i.e.~with neuronal activity traced backwards from target to start. (Right) The activity of the neuronal populations visualized with timing, demonstrating that threshold adaptation in combination with activity back-tracing enables path planning in \acp{snn}.}
  \vspace{-14pt}
  \label{fig:problem}
\end{figure}

While many models and learning rules for \acp{snn} are inspired by biology, an alternative approach is to adapt classical algorithms and data structures, which have shown to be efficient or optimal, to the unique architecture of \acp{snn} \cite{stagsted_towards_2020, lopez-randulfe_spiking_2021, aimone_provable_2021-1}. Although \acp{snn} are often recurrent, their connections are largely uni-directional with limited ability to propagate information backwards for network learning \cite{bellec2019SolutionLearningDilemma,lillicrap2020BackpropagationBrain,renner2024BackpropagationAlgorithmImplemented,maass1997NetworksSpikingNeurons,linnainmaa1970RepresentationCumulativeRounding,rumelhart1986LearningInternalRepresentations}. 
In addition to back-propagation, the concept of information back-tracing is a foundational technique to many graph and navigation algorithms that are based on Dijkstra’s algorithm~\cite{dijkstra1959NoteTwoProblems}. There is, however, no adequate implementation of back-tracing for \acp{snn} yet. 

Most existing graph-based \ac{snn} algorithms implement a backwards flow of information by altering the network structure or dynamics by introducing additional backward connections~\cite{roth_dynamic_1997, ponulak_rapid_2013} and plasticity rules \cite{schuman_shortest_2019}, respectively. Shortest path algorithms for \acp{snn} also commonly require the nodes (locations) in the graph to be represented by a single neuron \cite{ponulak_rapid_2013, schuman_shortest_2019, ruan_gsnn_2025}, even though most biological and neuromorphic navigation studies agree on the advantages of population codes for location representations \cite{hafting_microstructure_2005, dietrich_grid_2024, milford_ratslam:_2004, tang_real-time_2020}.

In this work, we investigated the issue of information back-tracing in a uni-directionally connected \ac{snn}, where a location is represented by a population code, with a focus on localization and navigation problems.
For this purpose, we interpret the abstract symbols introduced by the \ac{shtm} as places in the same way that conventional topological localization systems do \cite{xu2021topological,badino2011topometric,Maddern2012topological}. We then propose a novel approach to neuromorphic back-tracing that exploits the temporal dynamics of neurons in an \ac{snn} in the context of a simulated spatial navigation task, as shown in Fig.~\ref{fig:problem}.
Specifically, our contributions are:
\begin{enumerate}
    \item Introducing a \acf{stdta} for activity back-tracing in arbitrary uni-directional spiking neuron graphs through replay.
    \item Demonstrating the effectiveness of our approach in navigation tasks, achieving shortest path planning with fewer replays than steps from start to goal.
    \item Development and demonstration of an \acf{adta} method which identifies places with lower ambiguity, potentially enabling localization improvements after shortest path determination with \ac{stdta}.
    \item Code extensions to our openly accessible framework for the \ac{shtm} with code being publicly available at \url{https://github.com/dietriro/neuroseq}.
\end{enumerate}
\section{Related Work}

In this section, we review related work that covers the SNN model used in our simulations (Section \ref{sub:HTM}), neuromorphic navigation and localization (Section \ref{sub:neurolocal}), and finally path planning and back-tracing (Section \ref{sub:pathplan}).

\subsection{Hierarchical Temporal Memory Model}
\label{sub:HTM}
Our work is based on the \acf{htm} that models sequential processing in the neocortex of mammalian brains~\cite{hawkins_htm_2011}. We further draw inspiration from both, pre-play activity of neurons in the Hippocampus that predicts future paths \cite{pfeiffer2013HippocampalPlacecellSequences, moser2015PlaceCellsGrid,moser2017SpatialRepresentationHippocampal} and Theta phase precession, defined by unique temporal signatures of neuronal activity in the Hippocampus~\cite{okeefe_phase_1993, dragoi2006TemporalEncodingPlace}. The \ac{htm} uses multi-compartment neuron models with active dendrites in combination with \acf{sstdp} to learn sequences, predict future symbols, and replay already learned sequences~\cite{hawkins2016WhyNeuronsHave}. While it was shown to be effective in applications such as anomaly detection~\cite{wu_hierarchical_2018} and visual place recognition~\cite{neubert_neurologically_2019}, its abstract, binary implementation, however, does not include any neuronal dynamics close to biology.

\begin{figure}[!t]
  \centering
  \includegraphics[width=\columnwidth]{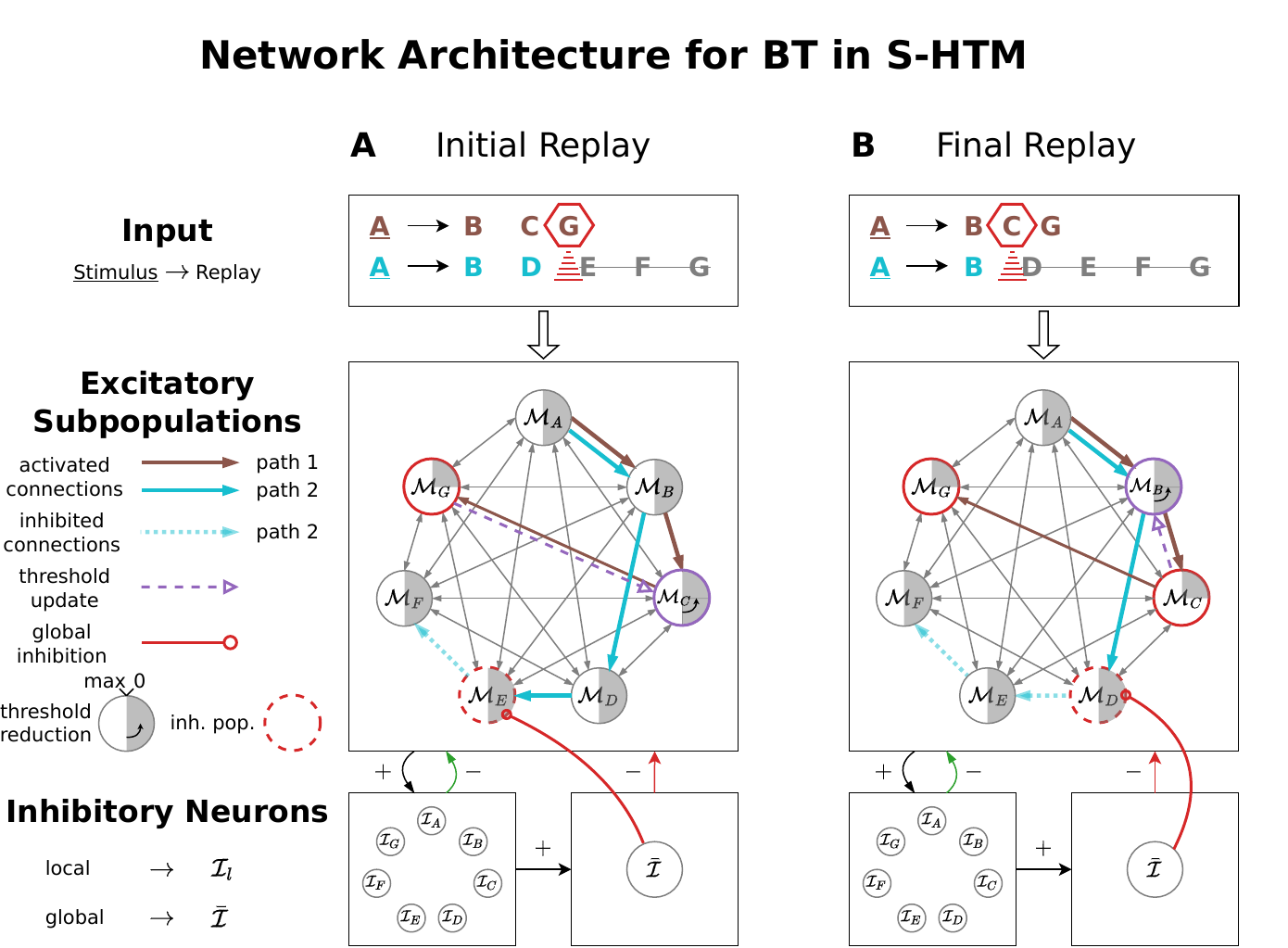}
  \caption{The network structure of our version of the \ac{shtm} after learning two sequences (brown, turquoise) from the environment shown in Fig.~\ref{fig:problem}.
  Akin to the original \ac{shtm} \cite{bouhadjar_sequence_2022}, our version consists of excitatory subpopulations \subpopm{l}, one for each location $l$. Each of these subpopulations is equipped with an inhibitory neuron $\mathcal{I}_l$, which is active during prediction and replay. The newly introduced global inhibitory neuron $\bar{\mathcal{I}}$ receives excitatory input from all local inhibitory neurons and maintains inhibitory connections to the excitatory subpopulations. It is only active during replay, to enable path selection by inhibiting alternative, slower paths.
  Adapted from \cite{bouhadjar_coherent_2023}.}
  \vspace{-15pt}
  \label{fig:network}
\end{figure}

The \ac{shtm} \cite{bouhadjar_sequence_2022}, a spiking version of the \ac{htm}, attempts to close this gap. Unlike the original \ac{htm}, which uses a distributed columnar structure, it organizes neurons into subpopulations (\subpopm{}), each representing a distinct symbol, such as \subpopm{A} for the start location (see Fig.~\ref{fig:network}). These subpopulations are managed by external and inhibitory neurons to control activation levels and ensure selective firing. The model implements spiking multi-compartment neurons that receive predictive inputs and trigger symbol onset, enabling the prediction of future states through connections learned by an \ac{sstdp} learning rule. By reducing the spike threshold of the excitatory neurons, the dendritic prediction alone is sufficient for causing a neuron to fire. With this mechanism, previously learned sequences can be replayed by triggering the first symbol of a sequence only (see Fig.~\ref{fig:network}). While the \ac{shtm} successfully replicates \ac{htm} functionality with enhanced biological plausibility, it remains primarily focused on biological feasibility rather than practical real-world or hardware applications, although its suitability for neuromorphic computing has already been demonstrated \cite{dietrich_sequence_2025-1, bouhadjar_sequence_2023}. Here, we use the \ac{shtm} in a shortest path planning task for localization and navigation for a real-world use case.

\subsection{Neuromorphic Localization and Navigation}
\label{sub:neurolocal}
Several neuromorphic systems have been proposed for localization and navigation. This includes spiking and non-spiking neural network architectures \cite{hwu2017neuromorphicnav,Mitchell2017neon,Dupeyroux2019,vanDijk2024}, insect-brain inspired approaches for route following and goal-directed homing tasks~\cite{schoepe2023FindingGoalInsectInspired,schoepe2023FindingGoalInsectInspired}, and more comprehensive approaches to localization that fuse neuromorphic sensors, hardware, and algorithms to perform visual place recognition to estimate an agent's position in the world~\cite{hines2024VPRTempoFastTemporally,wang2023BioinspiredPerceptionNavigation,yang2023NeuromorphicElectronicsRobotic,yu2023BraininspiredMultimodalHybrid,Gallego2020,kreiser2018path_integration,weikersdorfer2013simultaneous,Waniek2015Cooperative,dumont2023spiking}. 
Non-spiking neuromorphic approaches also demonstrate impressive performance across different functions, such as deployment on tiny drones for small-scale navigation and mapping or the aforementioned route following task \cite{Dupeyroux2019, vanDijk2024}. These results highlight the potential of neuromorphic principles and algorithms across a range of tasks, not limited just to navigation and localization.

\begin{figure*}
    \begin{subfigure}{.27\textwidth}
      \centering
      \includegraphics[width=\linewidth]{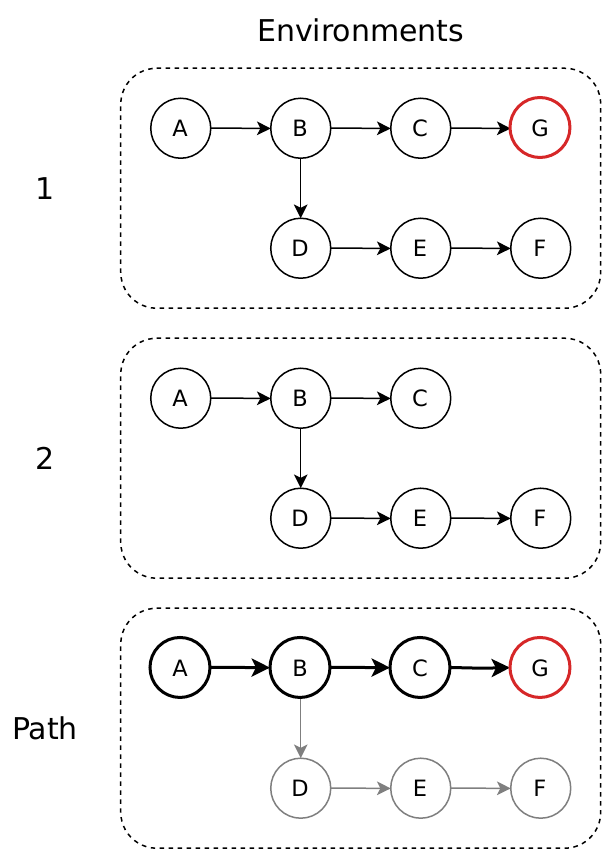}
      \caption{}
      \label{fig:backprop_process_a}
    \end{subfigure}%
    \hfill
    \begin{subfigure}{.65\textwidth}
      \centering
      \includegraphics[width=\linewidth]{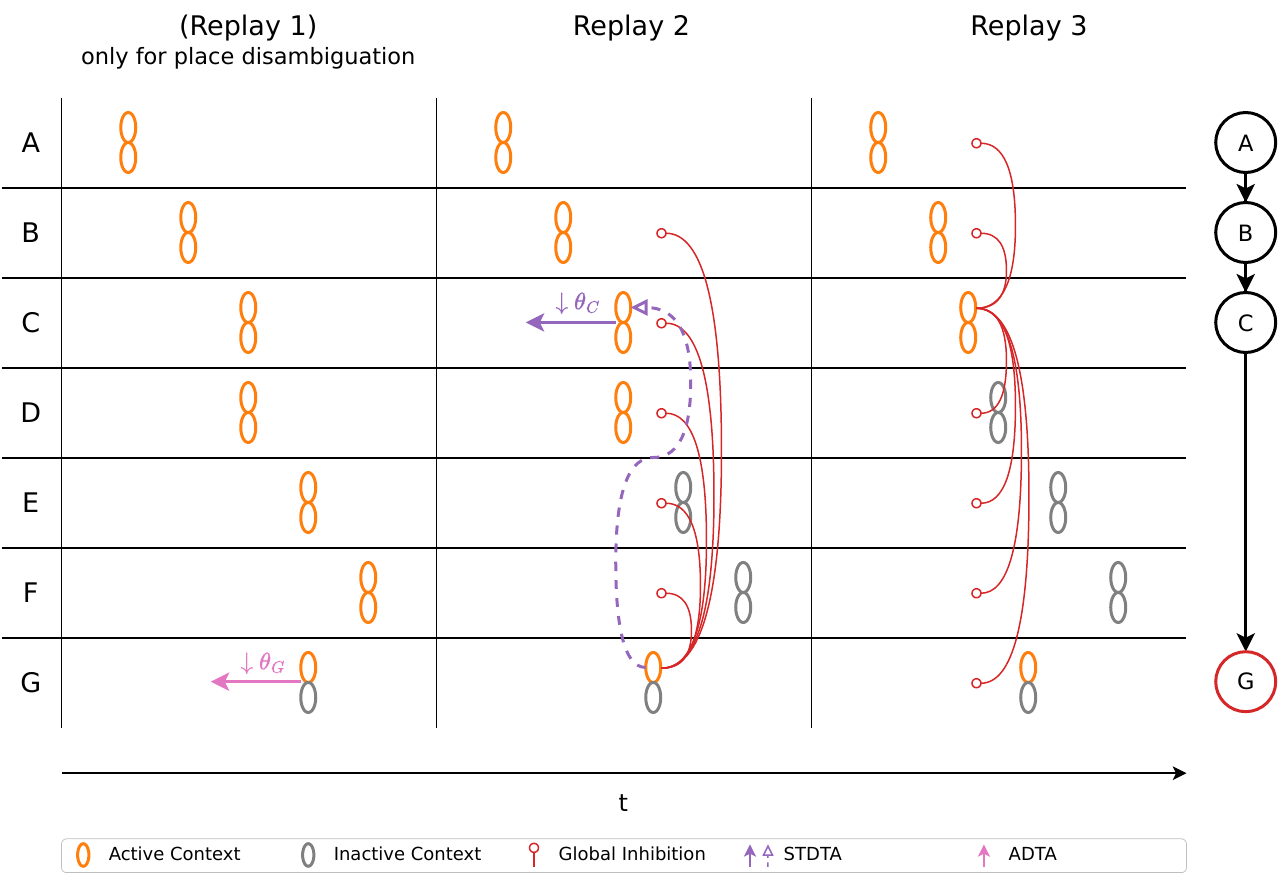}
      \caption{}
      \label{fig:backprop_process_b}
    \end{subfigure}
    \caption{(a) Two example environments alongside the target path. (b) An overview of 3 replay phases for a path planning or place disambiguation problem. The ovals represent an active (orange) or inactive (grey) subset of a subpopulation, representing a context. The first replay is only necessary for place disambiguation, where a target location with reduced ambiguity (less active neurons) is identified and its threshold is adapted (pink). During path planning, however, the target is manually set before Replay 2. Due to the reduced threshold, the neurons in subpopulation \subpopm{G} spike earlier than the neurons in \subpopm{E}, which are consequently inhibited by $\bar{I}$. After Replay 2, the information of the earlier spike time is propagated to the neurons in \subpopm{C} by a spike timing dependent rule, resulting in a threshold adaptation for \subpopm{C}. During the final replay, this leads to an inhibition of the competing subpopulation (path) \subpopm{D}, resulting in neuronal activity representing the final shortest path.}
    \vspace{-10pt}
    \label{fig:backprop_process}
\end{figure*}

\subsection{Path Planning and Back-Tracing}
\label{sub:pathplan}
 Numerous graph-based path planning and search algorithms use a variant of Dijstrak's algorithm, and therefore back-tracing, at their core~\cite{dijkstra1959NoteTwoProblems,hart1968FormalBasisHeuristic,stentz1994OptimalEfficientPath,goldberg2005ComputingShortestPath,dibbelt2014contractionhierarchies,bast2016routeplanning,funke2015provableefficiency,blum2019hardnessresults,haeupler2024UniversalOptimalityDijkstra,delling2009HighPerformanceMultiLevelGraphs}. In brief, these algorithms typically begin at a specific node in the graph and continue searching until the target node is found. Then, back-tracing traverses the graph in reverse order to trace a path from a target back to the start. This differs to back-propagation~\cite{linnainmaa1970RepresentationCumulativeRounding,rumelhart1986LearningInternalRepresentations} in that it does not carry gradient information but rather focuses on identifying specific sequences of nodes. The ability to perform back-tracing in \acp{snn} shows promise to be utilized in neuromorphic robotic localization and navigation tasks, where sequences of nodes correspond to recognizable places in the world, due to its inherently sequential process.

 To perform the back-tracing step of path planning in \acp{snn}, existing approaches often introduce additional backward connections, or explicitly reverse the direction of the computation~\cite{roth_dynamic_1997, ponulak_rapid_2013}. Alternatively, the shortest path can be determined by increasing connection weights during inference using plasticity rules \cite{schuman_shortest_2019} or entirely removing the weights of all other connections \cite{davies_advancing_2021}. Both approaches, however, permanently modify the network’s connections, thereby influencing potential future searches. Others used axonal delays based on environmental circumstances in combination with a list of spikes~\cite{krichmar2016PathPlanningUsing}, or generated tables of neuronal spike times for back-tracing~\cite{krichmar_flexible_2022, espino_rapid_2024}. These methods all share the need for additional mechanisms to determine the shortest path, whether through extra learning processes, delays, or backward connections. Neither of these methods shares our aim of enabling path planning through back-tracing in sequentially learned \acp{snn}, where a location is represented by multiple neurons.

\section{Methodology}
Our method builds upon the previously developed \ac{shtm} \cite{bouhadjar_sequence_2022}, which offers the ability to replay learned sequences. The \ac{shtm} model is capable of switching between a prediction mode, where it learns sequences of symbols, and a replay mode, where it is recalling or planning existing ones. The replay process is mitigated by adjusting activation thresholds of the neurons within the network after the prediction phase. This adjustment enables the network to activate neurons based solely on internal predictions, removing the need for external signals. A replay is then triggered by an external stimulus presented to the neuronal subpopulation representing the start of a sequence. A chain of activity then moves through the learned sequences of symbols until they all naturally conclude. This dynamic process is visualized in Figs.~\ref{fig:network} and \ref{fig:backprop_process}.

We will first discuss preliminaries for the implementation of our \ac{shtm} model in Section \ref{sec:prelims}, before introducing our approach for path planning by information back-tracing in a trained \ac{shtm} network with a pre-defined start and target location (Section \ref{sec:methods_shortest-path}). We then describe a modification of this approach, which solves the place disambiguation problem, i.e.~finding a less ambiguous or unique place for improving a location estimate (Section \ref{sec:methods_unique-place}). We use the example illustrated in Fig.~\ref{fig:backprop_process} throughout this section to explain both methods. 

\subsection{Preliminaries}
\label{sec:prelims}
In our localization and path planning task using the \ac{shtm}, each neuron subpopulation \subpopm{l} represents a specific location $l$ in a 2D environment (e.g.~$l \in \{A, B, C, ...\}$). We therefore also refer to these subpopulations as locations throughout this manuscript. These subpopulations are connected based on learned sequences \cite{bouhadjar_sequence_2022}. The trained network can be interpreted as a directed graph, where each node (neuronal subpopulation) corresponds to a place and the edges (connections) act as pathways between them (Fig.~\ref{fig:backprop_process_a}). This structure allows us to determine the shortest path by navigating through the network of spiking neurons, akin to finding the shortest path in a traditional graph. We use this sequence learning mechanism (prediction) of the original \ac{shtm} \cite{bouhadjar_sequence_2022} to generate these directed graphs of spiking neurons. Our path planning and place disambiguation algorithms are then built around the existing replay mechanism \cite{bouhadjar_sequence_2022}. 

The structure of the network is shown exemplary in Fig.~\ref{fig:network}. In addition to the local inhibitory neuron $I_l$ per excitatory subpopulation \subpopm{l} \cite{bouhadjar_sequence_2022}, we introduce an additional global inhibitory neuron $\bar{\mathcal{I}}$. This neuron receives excitatory connections from the local inhibitory neurons and maintains an inhibitory connection to each excitatory subpopulation. This way, only one global inhibitory spike is triggered for all subpopulations, even if multiple subpopulations are active concurrently, such as locations \textit{C} and \textit{D} in Fig.~\ref{fig:network}. The global inhibition is necessary to enable the subpopulations belonging to the shortest path to inhibit subpopulations belonging to alternative paths. 

\subsection{Shortest Path Finding in SNNs}
\label{sec:methods_shortest-path}
\textbf{Target selection:} The first step of our proposed method for shortest path finding in a sequentially connected \ac{shtm} network is the activation of a start subpopulation $\Omega$. Prior to triggering the first replay, however, we adapt the threshold $\theta_{\mathcal{M}_\Phi}$ of the membrane voltage before the first replay ($r_0$) for the entire excitatory subpopulation $\mathcal{M}_\Phi$ of the target location $\Phi$ as follows:
\begin{equation}
\label{eq:target-rule}
    \theta_{\mathcal{M}_\Phi}(r_0) = \theta_{\mathcal{M}_\Phi}(r_0) \cdot \lambda_\Phi,
\end{equation}
with $\lambda_{\Phi}$ defining the target threshold rate. In the example presented in Fig.~\ref{fig:backprop_process} the start population is defined as $\Omega = A$ and the target location as $\Phi = G$. The update (Eq. \ref{eq:target-rule}) in this example is then performed for the target \subpopm{G} before the first replay (Replay 2) of the path planning is initiated. In subsequent replays, the path is then back-traced through the subpopulations representing \textit{C} and \textit{B}, until the start location \textit{A} is reached. 

The replay process is initiated by an external stimulus presented to the start subpopulation \subpopm{A}. Due to the threshold reduction (Eq.~\ref{eq:target-rule}) applied to the target neuron subpopulation \subpopm{G} before the first replay, its firing time is shifted forward in time (Fig.~\ref{fig:backprop_process_b}, Replay 1, purple). This adjustment causes \subpopm{G} to inhibit neurons in \subpopm{E} through the global inhibitory neuron $\bar{\mathcal{I}}$, which otherwise would have fired concurrently with \subpopm{G} during Replay 2 (Fig.~\ref{fig:backprop_process_b}, red arrows). As a result, all alternative paths are suppressed by the global inhibition, i.e.~populations \subpopm{E} and \subpopm{F}. However, multiple paths remain active between populations \subpopm{C} and \subpopm{D} (Fig.~\ref{fig:backprop_process_b}), necessitating an additional mechanism to inhibit these intermediate, alternative paths.

\textbf{Back-tracing learning rule:} To address the limitation of unidirectional connections in \ac{shtm} networks, we introduce a method to back-trace path information without the need for additional backward connections. Instead of modifying the network architecture, we update the firing thresholds of the respective membrane potentials ($\theta_{\mathcal{M}_l}$) through a series of replays. This approach enables the network to inhibit alternative paths by performing a \acf{stdta} based on learned sequences.

After each replay iteration throughout the replay process, the threshold of a neuronal subpopulation \subpopm{m} is adapted if the spike time between that subpopulation and its connected subsequent subpopulation \subpopm{n} is within a certain range:  
\begin{equation}
    \Delta t_{\text{min}}^{b} < \Delta t(i, j) < \Delta t_{\text{max}}^{b},
\end{equation}
with $\Delta t_{\text{min}}^{b}$ and $\Delta t_{\text{max}}^{b}$ denoting the lower and upper bound, respectively, for the time difference $\Delta t(i, j)$ between a pre-synaptic spike of neuron $i \in \subpop{m}$ and a post-synaptic spike of neuron $j \in \subpop{n}$.  
The pre-synaptic subpopulation \subpopm{m} further needs to maintain a minimum number of connections to \subpopm{n} in order to activate the \ac{stdta}:
\begin{equation}
    N_{\text{con}}(i, j) > \rho,
\end{equation}
with $\rho$ denoting the targeted number of active neurons for a single context within a subpopulation. This rule assures that a pre-synaptic subpopulation $m$ is only considered for an update if it maintains enough connections to the post-synaptic subpopulation $n$ to have potentially triggered its activation.

Finally, the threshold for each subpopulation \subpopm{l} satisfying both of these constraints is updated after every replay $r$: 
\begin{equation}
\label{eq:trace-rule}
    \theta_{\mathcal{M}_l}(r) = \theta_{\mathcal{M}_l}(r-1) \cdot \lambda_b,
\end{equation}
with $\theta_{\mathcal{M}_l}(r-1)$ denoting the value $\theta_{\mathcal{M}_l}$ at the previous replay ($r-1$) and $\lambda_b$ defining the back-tracing rate. 

In the example depicted in Fig.~\ref{fig:backprop_process_b}, the threshold for subpopulation \subpopm{C} is thus decreased because it maintains connections to \subpopm{G} and the spike time difference $\Delta t(C, G)$ between \subpopm{C} and \subpopm{G} is lower due to the reduced threshold of \subpopm{G}. Conversely, the threshold for subpopulation \subpopm{D} remains unchanged, since it does not maintain any connections to \subpopm{G} and no other populations were active after \subpopm{D} during the second replay. This selective threshold adjustment based on spike times effectively inhibits non-optimal paths, ensuring that only the shortest path in the network is active. 

This process ultimately results in a single, shortest path from the start to the target subpopulation being active during the final replay (see Fig.~\ref{fig:backprop_process_b}, Replay 3). In this step, the reduced threshold (Replay 2, purple arrow) causes neurons in \subpopm{C} to spike earlier, thereby inhibiting the neurons in \subpopm{D} (Replay 3, red arrows). In this toy example, the back-tracing process required only one replay step. However, in larger environments, this process may require multiple replay steps, equal to the number of intermediate locations with overlapping active populations (e.g., \subpopm{C} and \subpopm{D} in this example).

\subsection{Place Disambiguation in SNNs}
\label{sec:methods_unique-place}

\textbf{Population encoding for unique places:} In the previous section, we described a solution to the general path planning problem with a pre-defined start and goal location. There are, however, cases, where the goal location is not known \emph{a priori}, such as in the case of localization. When a person or an agent is placed in a new environment at an ambiguous location, such as the beginning of a hallway that looks the same on many floors, finding a less ambiguous location to localize oneself becomes a crucial task. An example for such environments is shown in Fig.~\ref{fig:backprop_process_a} (Top). The two environments visualized here are almost identical, besides location $G$, which is only present in Environment 1. Since location G can only occur in a single context, a smaller subset of neurons from subpopulation \subpopm{G} would be active during a replay, compared to the other, aliased locations (e.g.~\subpopm{E}). It is therefore considered advantageous to visit location $G$ in order to improve the location estimation of the agent and distinguish between the Environments~1~and~2. 

\textbf{Ambiguity dependent threshold adaptation:} To distinguish ambiguous from more unique places, we propose a rule for adaptive neural thresholding based on the ambiguity of a place, i.e.~the activity of the entire subpopulation. We define this rule as an \acf{adta}. A subpopulation of neurons representing a symbol can constitute multiple different contexts. The number of contexts each subpopulation can represent is determined mainly by the target number of neurons per context ($\rho$) and the total number of neurons within the subpopulation ($N$). After learning,  the ambiguity is hence a direct correlation between the number of active neurons within a subpopulation and the number of contexts represented by these neurons. We use this property of the network to define the threshold update for all neurons in subpopulation \subpopm{l} as follows:
\begin{equation}
\label{eq:loc_target}
    \theta_{\mathcal{M}_l}(r) = \theta_{\mathcal{M}_l}(r) \cdot e^{ \gamma (F_{a} - F_{\rho})} \cdot \lambda_a,
\end{equation}
with
\begin{equation}
\setstretch{1.5}
\label{eq:loc_target_b}
    \begin{matrix}
        F_{a} = \frac{N^{\text{act}}_{l}}{N_{l}}~ \text{and}\ F_{\rho} = \frac{\rho}{N_{l}}, \\    
    \end{matrix}
\end{equation}
where $F_{a}$ and $F_{\rho}$ define the fraction of active and targeted (per context) neurons, respectively. The threshold adaptation rate based on ambiguity is defined by the variable $\lambda_a$. The total number of excitatory neurons in \subpopm{l} is defined by the parameter $N_l$. The number of active neurons in \subpopm{l} is defined by $N^{\text{act}}_l$ and determined by grouping the spikes within the respective subpopulations for each replay individually. 
The slope of the exponential increase or decay is defined by $\gamma$ as follows:
\begin{equation}
\setstretch{1.5}
\label{eq:loc_target_c}
    \begin{matrix}
        \gamma = 
        \begin{cases}
            \ \gamma^+ & \text{if}\ F_{a} \geq F_{\rho},\\
            \ \gamma^- & \text{else.}
        \end{cases}    
    
    \end{matrix}
\end{equation}

This threshold adaptation is applied after the \ac{stdta} update and results in an additional shift of the firing time for neurons relative to the fraction of active neurons in a subpopulation. This is shown for Replay 1 in Fig.~\ref{fig:backprop_process_b}, where only one group of neurons (one context) is active for \subpopm{G} opposed to all other subpopulations, maintaining two active contexts each. The threshold for \subpopm{G} is thus reduced based on the \ac{adta} rule (pink arrow). Due to the reduced threshold, the neurons in \subpopm{G} subsequently fire earlier during Replay 2, akin to the behavior of the same population in the previous path planning scenario. This consequently results in an inhibition of the other subpopulations (\subpopm{E}) by the global inhibitory neuron $\bar{\mathcal{I}}$, preventing them from firing at all (red arrows). After this step, the same process as described in Section \ref{sec:methods_shortest-path} is carried out to identify the shortest path to this new, unambiguous target.

\section{Experimental Results}

In this section, we introduce the experimental setup, the software and parameters used for our evaluations (Section \ref{sec:eval_setup}). The experiments presented thereafter are separated into two parts. First, we evaluate the path planning capabilities with a specified target in Section \ref{sec:eval_path}. Subsequently, we analyze the performance of the network for place disambiguation in Section \ref{sec:eval_loc}.

\subsection{Experimental Setup}
\label{sec:eval_setup}

The presented algorithm was developed and implemented in the spiking network framework PyNN \cite{davison2009pynn}, with the NEST simulator as a backend \cite{Gewaltig:NEST}, and is built on previous work \cite{bouhadjar_sequence_2022, dietrich_sequence_2025-1}. The general process we use for performing the replay as well as most of the parameters are the same as presented in earlier studies \cite{bouhadjar_sequence_2022, bouhadjar_coherent_2023}. Deviations from the process or parameters are mentioned in the following and throughout the remaining evaluation where applicable. The parameter changes are summarized in Table~\ref{tab:parameters}. We set the random seed responsible for, e.g, sampling the connections between subpopulations to a constant value of 5 for all experiments.

\begin{figure*}[t]
    \begin{minipage}[b]{0.26\textwidth}
        \begin{subfigure}{\textwidth}
          \centering
          \includegraphics[width=\linewidth]{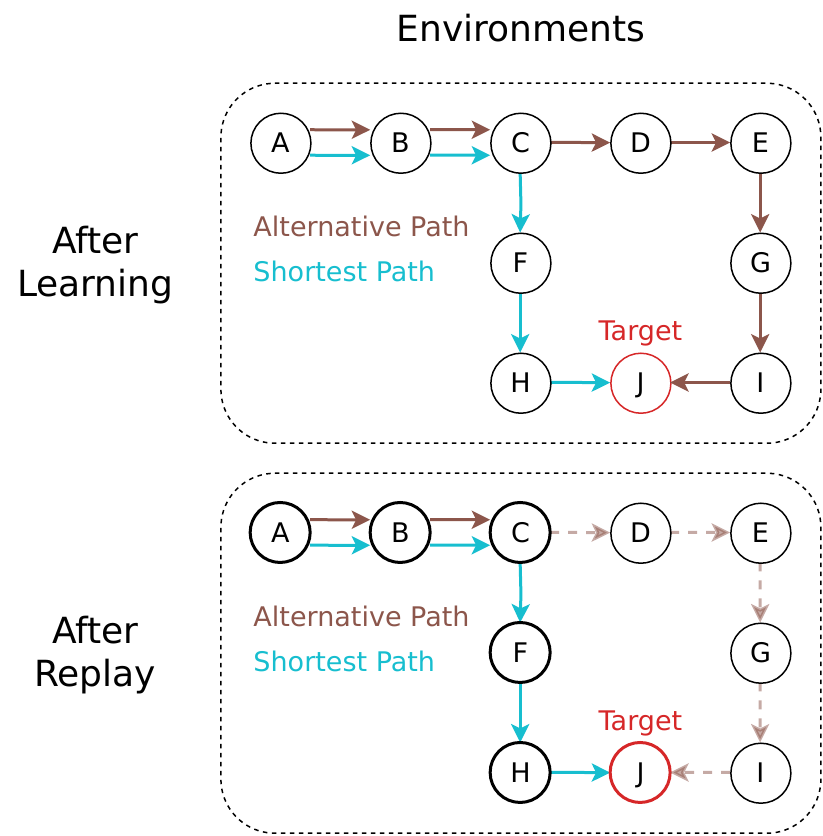}
          \caption{}
          \label{fig:eval_path_a}
        \end{subfigure}%
        \\[\baselineskip]
        \begin{subfigure}{\textwidth}
          \centering
          \includegraphics[width=\linewidth]{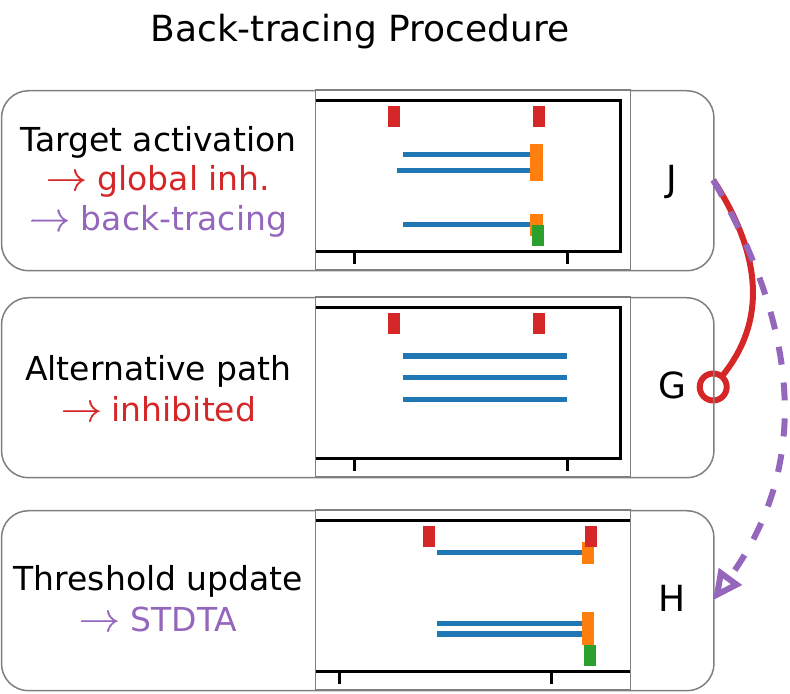}
          \caption{}
          \label{fig:eval_path_b}
        \end{subfigure}%
\end{minipage}
\hfill
\begin{subfigure}[b]{.67\textwidth}
  \centering
  \includegraphics[width=\linewidth]{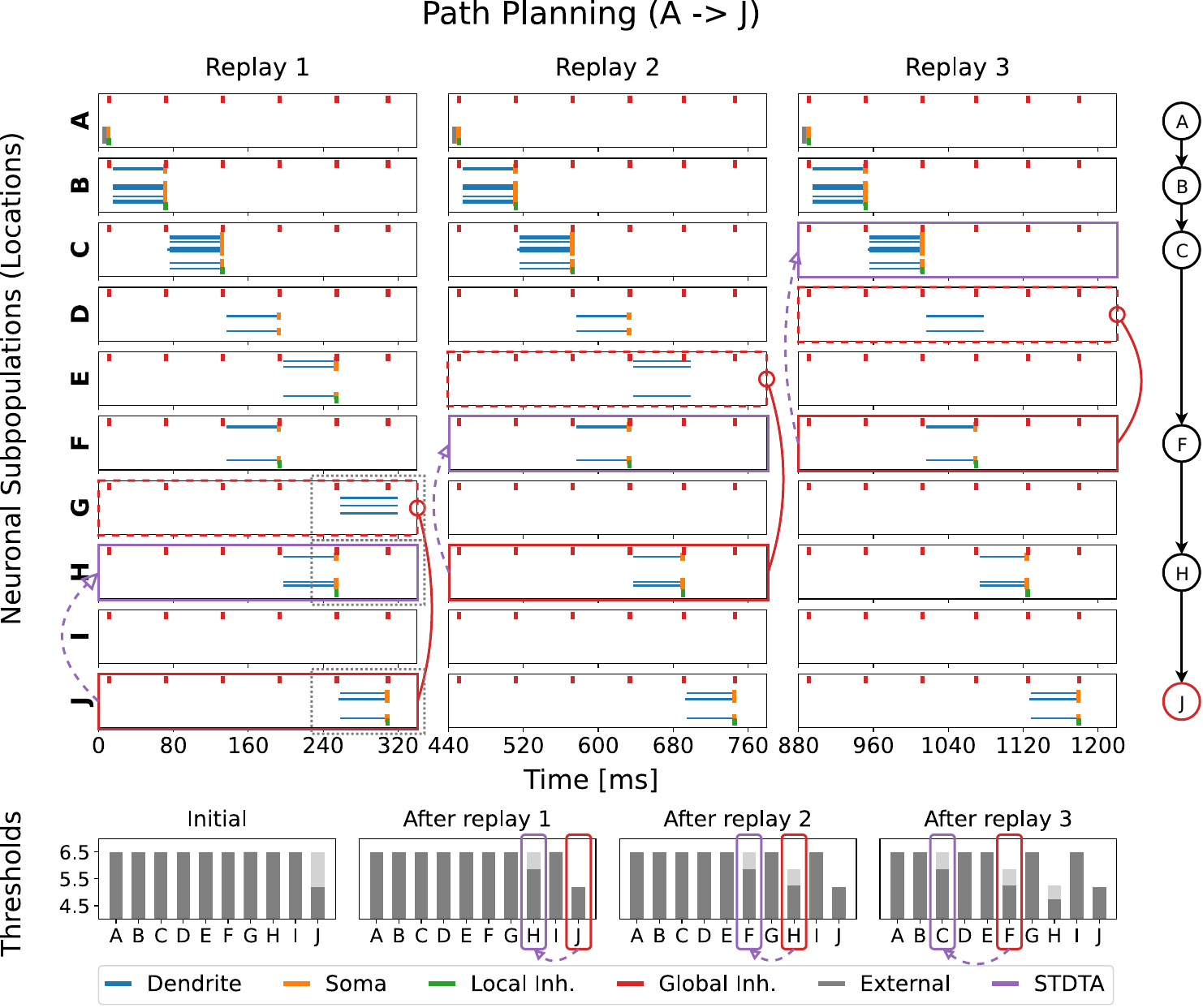}
  \caption{}
  \label{fig:eval_path_c}
\end{subfigure}
\caption{(a) (Top) A graphical representation of active connections between locations before and (bottom) after replay with the shortest path (turquoise), an alternative path (brown), and the target (red). (b) An overview of the corner stones of the back-tracing and threshold adaptation process, showing events from populations \subpopm{J}, \subpopm{G}, and \subpopm{H}. (c) (top) The results of the path planning replay process for a target path from \textit{A} to \textit{J} with the events for all neuronal populations, the final path of active neuronal populations (right), (bottom) and the thresholds throughout the replay phases. The neuronal events include a somatic spike (orange), triggered by a dendritic plateau potential (blue), or the initial, external input (grey), together with local (green) and global inhibition (red). The steps of the back-tracing are highlighted for each phase, including the current target (solid red), the inhibited alternative population (dashed red), and the populations performing an STDTA update (purple).}
\vspace{-15pt}
\label{fig:eval_path}
\end{figure*}

We define the ambiguity $\alpha$ of a location $l$ as the number of occurrences in a set of partially overlapping environments:
\begin{equation}
    \alpha (l) = o(l, E),
\end{equation}
where $o(l, E)$ defines the number of occurrences of location $l$ in a set of environments $E$. A location $l_1$ which occurs once in a set of environments therefore has the lowest ambiguity value of $\alpha(l_1)=1$ while a place $l_2$ occurring in $x$ of the given environments has a value of $\alpha(l_2) = x$. This value is also directly proportional to the number of neurons active for a place during replay:
\begin{equation}
    \bar{N}_{l}^{act} = \alpha(l) \cdot \rho,
\end{equation}
where $\bar{N}_{l}^{act}$ is the targeted number of active neurons per subpopulation defined in the parameters. 

For the threshold update in Eq.~\ref{eq:loc_target}, we set $\gamma^+=-8$ and $\gamma^-=20$ for all experiments. We identified suitable values for $\gamma$ by hand, which yielded good results for the examples described in this section, the value of which would change with different required network parameters, especially the maximum target offset. The upper threshold for the back-tracing update $\tmaxb$ was selected to guarantee that only subpopulations with active connections to a post-synaptic subpopulation which already has a reduced threshold are updated. This value fluctuates based on the environments, connections, and initializations and is therefore slightly different in the path-planning experiment than in the ambiguity experiments. We increase this value for the \textit{ambiguity-02b} experiment to explicitly allow the update for all active subpopulations.  

\begin{table}[b]
\vspace{-10pt}

    \caption{The parameters used in each of the path planning and place disambiguation experiments.}
    \centering
    \small
    \begin{tabular}{l|c|c|c|c|c|c}
        Experiment & $N$ & $\rho$ & $\Delta t_{\text{max}}^{b}$ & $\lambda_\Phi$ & $\lambda_b$ & $\lambda_a$\\
        \hline
        path-planning (Fig.~\ref{fig:eval_path}) & 21 & 3 & 58 & 0.8 & 0.9 &  -  \\
        ambiguity-01  (Fig.~\ref{fig:eval_amb_01}) & 21 & 3 & 55 &  -  & 0.9 & 0.2 \\
        ambiguity-02a (Fig.~\ref{fig:eval_amb_02_a}) & 21 & 3 & 55 &  -  & 0.9 & 0.2 \\
        ambiguity-02b (Fig.~\ref{fig:eval_amb_02_b}) & 21 & 3 & 60 &  -  & 0.9 & 0.2 \\
    \end{tabular}
    \label{tab:parameters}
\end{table}

All networks used in our experiments have been trained for 50 epochs on a set of sequences in the same way as presented in previous works \cite{bouhadjar_sequence_2022, bouhadjar_coherent_2023}. The start location for all sequences is chosen to be \textit{A}. We assume that, during training, the start location as well as the onset of a new sequence is known. Hence, we activate a sparse, sequence specific firing of the start subpopulation \cite{bouhadjar_sequence_2022}. 
When switching from training to replay mode, we once set the neuronal threshold for all populations $l$ to $\theta_{\mathcal{M}_l} = 6.5$ to enable somatic spikes being triggered solely by dendritic input \cite{bouhadjar_sequence_2022}.
We then perform a number of replays until the network converges to a solution. There is currently exists no automated stop condition for the replay process.

\subsection{Shortest Path Finding in SNNs}
\label{sec:eval_path}

\begin{figure*}[!t]
\hfill
\begin{subfigure}{.15\textwidth}
  \centering
  \includegraphics[width=\linewidth]{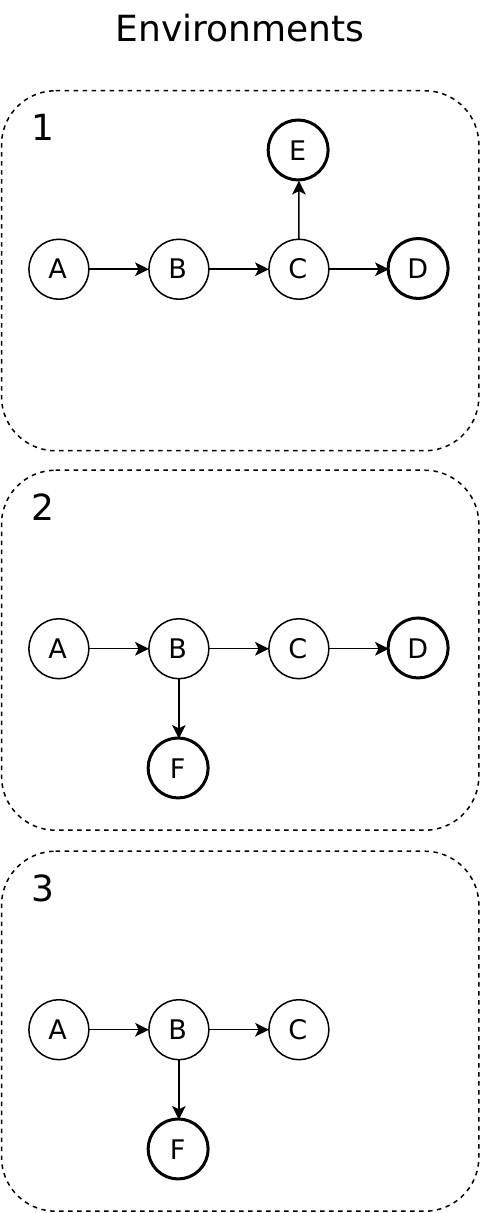}
  \caption{}
  \label{fig:eval_amb_01_a}
\end{subfigure}%
\hfill
\begin{subfigure}{.45\textwidth}
  \centering
  \includegraphics[width=\linewidth]{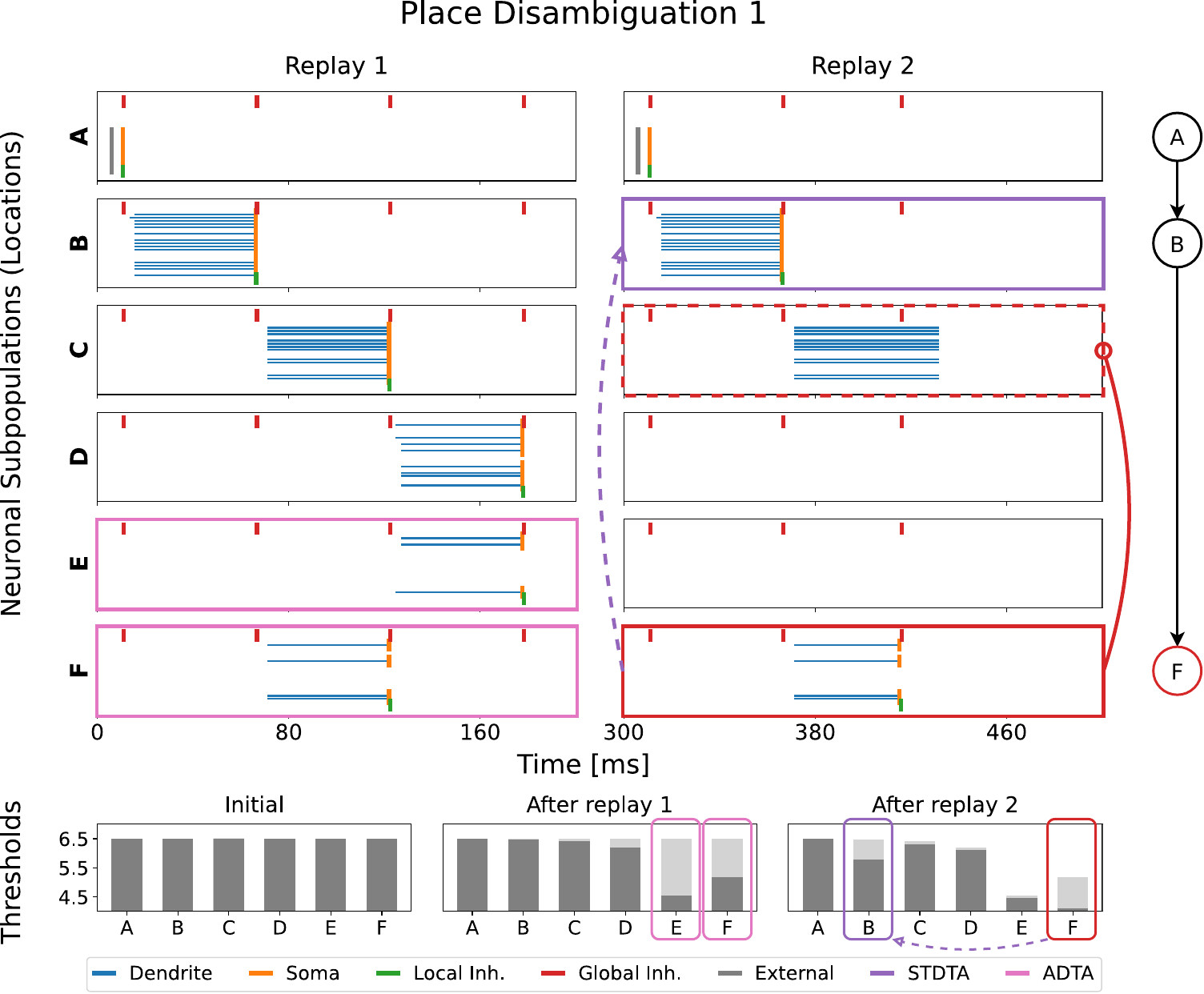}
  \caption{}
  \label{fig:eval_amb_01_b}
\end{subfigure}
\hfill
\caption{(a) A graphical representation of the environments used for the place disambiguation experiments, environments 1 and 2 are used for the first experiment and all three for the second. (b) The results of the first place disambiguation experiment for environments 1 and 2, with the events for all neuronal populations (top), the final path of active neuronal populations (right), and the thresholds throughout the replay phases (bottom). 
The neuronal events are depicted as in Fig.~\ref{fig:eval_path_c}, except for the newly introduced \ac{adta} updates due to reduced ambiguity (pink).}
\vspace{-15pt}
\label{fig:eval_amb_01}
\end{figure*}

We evaluated the path planning capabilities of the \ac{shtm} on a single environment which is shown in Fig.~\ref{fig:eval_path_a}. The environment consists of 10 unique locations. The network is trained on two sequences ([\textit{A, B, C, F, H, J}], [\textit{A, B, C, D, E, G, I, J}]), where the first one constitutes the shortest path to the target \textit{J} and the second one a longer, alternative path. 

The results, shown in Fig.~\ref{fig:eval_path_b}, demonstrate that the network is capable of calculating the path from the start location \textit{A} to the target location \textit{J} within three iterations of replay. During the first replay, the active neurons in subpopulation \textit{J} fire earlier compared to the ones in \textit{G}, due to the reduced threshold of the neurons in \textit{J} for being in the target subpopulation. The neurons in \subpopm{J} therefore trigger a globally active inhibition (red) and prevent \subpopm{G} from firing. 
The reduced threshold of \subpopm{J} consequently leads to an earlier spike time of \subpopm{J} with respect to the spike time of the previous location \subpopm{H}.

Based on the learning rule presented in Section \ref{sec:methods_shortest-path}, the threshold for all neurons in \subpopm{H} gets reduced according to Eq.~\ref{eq:trace-rule}. Therefore, this subpopulation of neurons inhibits the neurons in \subpopm{E} during the second replay. This two-step process is visualized separately in the zoomed clippings in Fig.~\ref{fig:eval_path_b} for the inhibition triggering, earlier spiking population \subpopm{J} (top), the inhibited population \subpopm{G} (center), and the subpopulation receiving a threshold update through \ac{stdta} \subpopm{H} (bottom). 

The same mechanism is applied to the subpopulation \subpopm{F}, which then inhibits \subpopm{D} during the final (third) replay. The activity is successfully propagated backwards to the first common location \textit{C}, leaving only one path of subpopulations to the target $J$ active, the shortest path. As demonstrated by this experiment, the number of replays required for an arbitrary path calculation is always equal to the number of locations (subpopulations) within a shortest path where another subpopulation is active as well, i.e.~the number of simultaneously active locations. Therefore, in most scenarios, much less replays are required than the number of total steps to a goal, since there is usually an overlap between paths.

\subsection{Place Disambiguation in SNNs}
\label{sec:eval_loc}

In the second part of our experiments, we evaluated the ability of the presented method to identify unique or less ambiguous places in two different scenarios. The first scenario uses two very similar environments to demonstrate the general capability of the algorithm to find unique places in environments with many ambiguous places and subsequently identify the shortest path to it. The second scenario adds another environment to analyze the behavior of our learning algorithm when multiple levels of ambiguity exist among the different places in the environments. The sequences used for training the networks include all possible paths from the start location \textit{A} to any final location in the respective environments, e.g.~\textit{E} and \textit{D} in the first environment (Fig.~\ref{fig:eval_amb_01_a}). 

\textbf{Binary disambiguation:} The first two environments share the center part (\textit{A, B, C, D}), while both include one unique location (\textit{E} and \textit{F} respectively). The unique location in the first environment (\textit{E}), however, is one hop further away from the start location \textit{A}. We would therefore expect the network to prefer location \textit{F} over \textit{E}. The results for these experiments are shown in Fig.~\ref{fig:eval_amb_01}. They demonstrate that the network identifies the two unique places \textit{E} and \textit{F} during the first replay and reduces their threshold thereafter (pink) based on Eq.~\ref{eq:loc_target}. Since \subpopm{F}, however, spikes earlier than \subpopm{E}, it inhibits population \subpopm{C}, which leads to the alternative unique place. Therefore, while the firing thresholds of the neurons in both populations are reduced, only the one that is closest to the start population in the number of steps wins. In this example, the activation does not need to be back-traced further as this initial inhibition of \subpopm{C} during the second replay is sufficient for identifying a single path to the closest unique place \textit{F} (Fig.~\ref{fig:eval_amb_01_b}, right). 

\textbf{Multiple ambiguities:} In our second place disambiguation experiment, we demonstrate not only the back-tracing of the path to the goal but also analyze the replay behavior in the presence of multiple places with varying ambiguity. For this purpose, we added Environment 3 to our world, which shares the previously unique place \textit{F} with Environment 2. The only unique place remains \textit{E} ($\alpha(E)=1$), while \textit{F} and \textit{D} become more ambiguous ($\alpha(F) = \alpha(D) = 2$) in this new scenario, ranking now between place \textit{E} and the center part (\textit{A, B, C} with $\alpha=3$). 

\begin{figure*}
\begin{subfigure}{.49\textwidth}
  \centering
  \includegraphics[width=\linewidth]{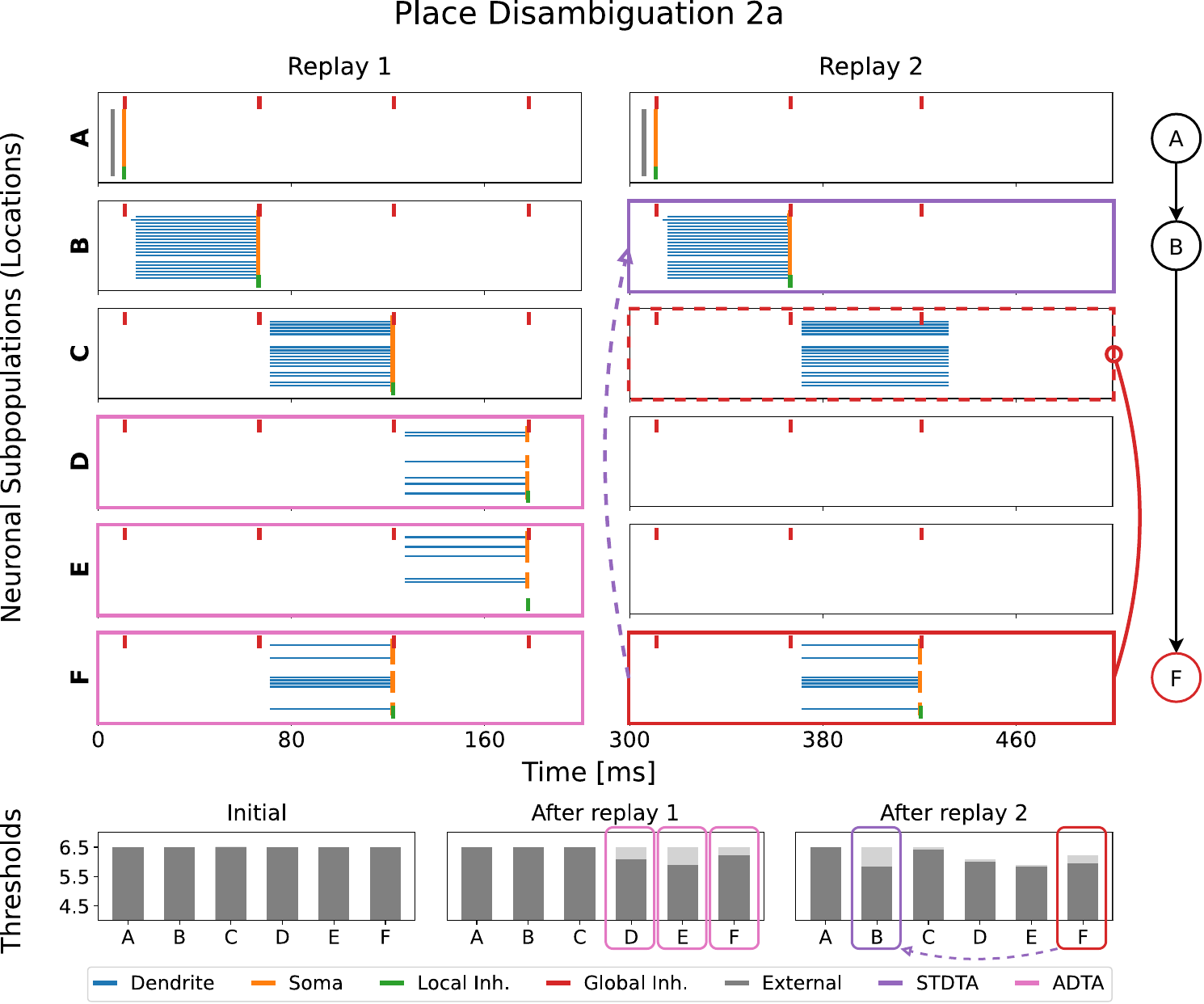}
  \caption{}
  \label{fig:eval_amb_02_a}
\end{subfigure}%
\hfill
\begin{subfigure}{.49\textwidth}
  \centering
  \includegraphics[width=\linewidth]{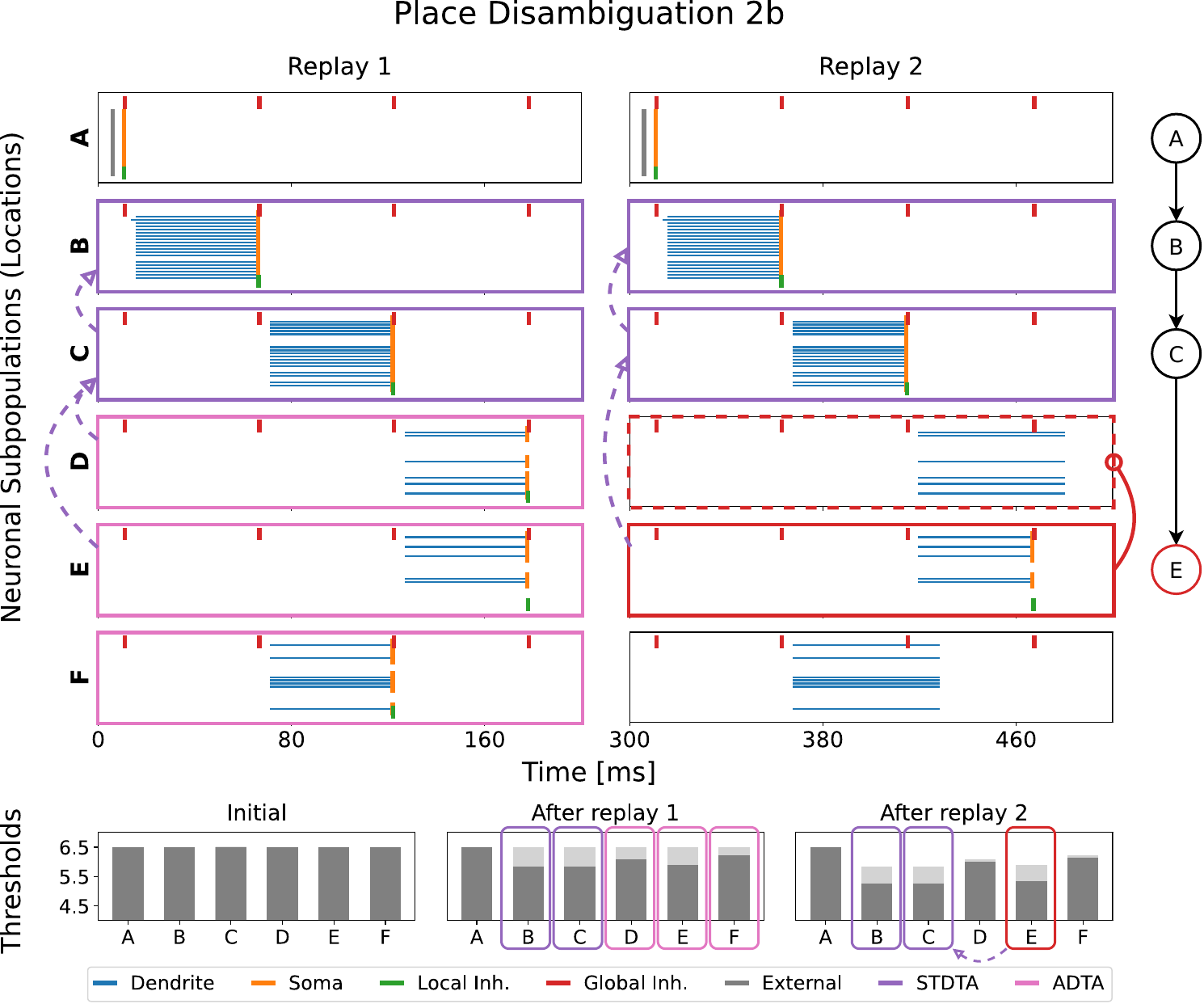}
  \caption{}
  \label{fig:eval_amb_02_b}
\end{subfigure}
\caption{The results of the second place disambiguation experiment, including all three environments (Fig.~\ref{fig:eval_amb_01_a}). Two separate runs, \textit{a} and \textit{b}, are shown with different parameters, one is favoring the closest place with reduced ambiguity (a), the other favors the place with minimal ambiguity but higher distance to the start location. For each experiment, the events for all neuronal populations are shown (top) and the thresholds throughout the replay phases (bottom) are visualized. The neuronal events are depicted as in Fig.~\ref{fig:eval_amb_01_b}.}
\vspace{-10pt}
\label{fig:eval_amb_02}
\end{figure*}

Depending on the target application there can be two desired outcomes for this scenario. Either the closest place with reduced ambiguity is targeted (\textit{F}) or the place with the least ambiguity is targeted regardless of the distance (\textit{E}), therefore place \textit{D} should never be selected as the target. The default behavior of the network, using the parameter configuration of the previous experiments, is shown in Fig.~\ref{fig:eval_amb_02_a}. It results in the first behavior, as it always targets the closest place that is less ambiguous than other equidistant places. In this experiment, the upper spike timing threshold of the back-tracing $\Delta t_{\text{max}}^{b}$ is chosen to be small enough to only allow updates for neurons, whose post-synaptic neuron already has a reduced threshold, i.e.~an earlier spike time leading to a $\Delta t (i, j) < \Delta t_{\text{max}}^{b}$. 

Opposed to this, an increase of this parameter eventually enables a spike threshold update (\ac{stdta}) for each subpopulation that precedes another active subpopulation during replay. This potentially enables subpopulations to inhibit other subpopulations representing locations with less but not the least ambiguity, such as the location \textit{F}. We tested this behavior on the given environments with $\Delta t_{\text{max}}^{b} = 60$. The results, visualized in Fig.~\ref{fig:eval_amb_02_b}, show that, although the threshold for \subpopm{F} is reduced after the first replay, the overall threshold adaptation outweighs this reduction. This allows the back-tracing of the path to the least ambiguous location $E$ within the next round of replay. 

This behavior is governed by the upper spike timing threshold of the back-tracing $\Delta t_{\text{max}}^{b}$, the back-tracing rate $\lambda_b$, the ambiguity dependent threshold adaptation rate $\lambda_a$, and the slope of the exponential increase/decay defined by $\gamma$ (see Eq.~\ref{eq:loc_target}). Each of these parameters influences the behavior in its own way. Suitable parameter combinations therefore have to be found individually for different situations.

\section{Conclusion and Future Work}
\label{sec:discussion}
In this paper, we have presented a novel technique for shortest path finding and place disambiguation in spiking neural networks applied to navigational tasks.
Through \acf{stdta}, our network is capable of back-tracing activity from a target to a start population of neurons by modulating the firing activity (threshold) of the neurons. We demonstrated that this mechanism is capable of effectively inhibiting alternative, longer paths during a set of replays. We further showed that the number of replays required is generally limited by the number of places in the shortest path.
Beyond that, we introduced a novel method for identifying places with reduced ambiguity, which shows promise for improving the localization estimate of an agent. We showed that, by using an \acf{adta} rule, we can identify places with reduced ambiguity and distinguish between places with different levels of ambiguity. 

We thereby laid the foundation for future applications of the \ac{shtm} or similar networks to real-world navigation and localization problems. Whilst we simulate input data in this study as a proof-of-concept for effective shortest path planning, future work will focus on integrating and processing real-world sensor data with the \ac{shtm}. This will allow more complex experiments and benchmarks with other localization and navigation systems, such as simultaneous localization and mapping (SLAM)~\cite{yang2023NeuromorphicElectronicsRobotic,weikersdorfer2013simultaneous}. 

The design of our proposed replay learning mechanism will also allow a more efficient implementation on a range of neuromorphic processors, since the threshold adaptation requires only local learning. The spiking path planning and place disambiguation could therefore be deployed as a neuromorphic navigation system on processors such as Loihi \cite{davies2018LoihiNeuromorphicManycore}, Speck \cite{Yao2024speck}, or SpiNNaker \cite{furber2014SpiNNakerProjecta}. This fusion of neuromorphic software and hardware would further allow for application-oriented benchmarks of the back-tracing with respect to energy efficiency and inference time.

Beyond navigation, our method provides a useful means of learning and processing sequential information, with potential benefits in a variety of applications, such as natural language processing \cite{Pandey2023} or anomaly detection \cite{wu_hierarchical_2018}. 
From a theoretical perspective, we will further analyze the timing protocol used for threshold adaptation and generalize it beyond \ac{shtm}, while drawing on insights from the Transition-Scale-Space model that accelerates the retrieval of place-sequences in navigation~\cite{storesund2024SimulatingTransitionCell}.

In conclusion, we have demonstrated the effectiveness of replay and threshold adaptation for path planning and localization in a functional real-world use case, showcasing the benefit of using biologically inspired neural networks, such as \acp{snn}, to perform complex sequence-based tasks in localization and navigation.

\balance

\printbibliography

\end{document}